This paper has been accepted for publication and is forthcoming in *International Journal of Human-Computer Interaction*.

# Trust in AI and Its Role in the Acceptance of AI Technologies

Hyesun Choung[1]*

Prabu David[2]

Arun Ross[3]

*Corresponding author, choungh@msu.edu

[1] College of Communication Arts and Science, Michigan State University
[2] Department of Media and Information, Department of Communication, Michigan State University
[3] Department of Computer Science and Engineering, Michigan State University

## Trust in AI and Its Role in the Acceptance of AI Technologies

### Abstract

As AI-enhanced technologies become common in a variety of domains, there is an increasing need to define and examine the trust that users have in such technologies. Given the progress in the development of AI, a correspondingly sophisticated understanding of trust in the technology is required. This paper addresses this need by explaining the role of trust on the intention to use AI technologies. Study 1 examined the role of trust in the use of AI voice assistants based on survey responses from college students. A path analysis confirmed that trust had a significant effect on the intention to use AI, which operated through perceived usefulness and participants' attitude toward voice assistants. In Study 2, using data from a representative sample of the U.S. population, different dimensions of trust were examined using exploratory factor analysis, which yielded two dimensions: human-like trust and functionality trust. The results of the path analyses from Study 1 were replicated in Study 2, confirming the indirect effect of trust and the effects of perceived usefulness, ease of use, and attitude on intention to use. Further, both dimensions of trust shared a similar pattern of effects within the model, with functionality-related trust exhibiting a greater total impact on usage intention than human-like trust. Overall, the role of trust in the acceptance of AI technologies was significant across both studies. This research contributes to the advancement and application of the TAM in AI-related applications and offers a multidimensional measure of trust that can be utilized in the future study of trustworthy AI.

## Trust in AI and its Role in the Acceptance of AI Technologies

Inspired by human intelligence, artificial intelligence (AI) systems are becoming increasingly adept in their ability to learn, reason, self-correct and emulate human decisions in some domains (Russell et al., 2016; Watson, 2019). AI is ubiquitous in modern life, enabling smart technologies and applications such as smart home devices, fitness trackers, autonomous driving systems, and social media platforms (Gorwa et al., 2020; Lockey et al., 2021). Further, AI technologies are equipped with varying degrees of autonomy that minimize the need for human control or oversight, and many AI-enabled devices are imbued with anthropomorphic features and natural language processing capabilities, turning them into social actors (Watson, 2019). The pervasiveness and impact of AI has raised concerns about its ethics and principles of governance.

Experts contend that without guiding principles the application of AI in critical areas such as health, finances, and criminal justice will perpetuate the same biases in human thinking, thus robbing the technology of its potential to create new systems that enlighten and augment human intelligence to correct social wrongs (AI HLEG, 2019). Also, there is a growing fear that our reliance on AI comes at the cost of human agency, and that fear is further stoked by the black-box nature of machine learning algorithms, which generate a response that mimics human response without adequate explanation of the underlying process (Barredo Arrieta et al., 2020).

Such concerns, compounded by the erosion of trust in institutions and governments (Edelman, 2021), underscore the importance of building trustworthy AI system to instill confidence among users (Gillath et al., 2021; Thiebes et al., 2020) and to further our understanding of the social and psychological mechanisms of trust in human-AI interaction (Gillath et al., 2021). Extant literature on trust in AI and algorithm-driven technologies has focused on exploring the key principles of trustworthy AI systems such as beneficence, non-maleficence, autonomy, justice, and explicability (Thiebes et al., 2020). Some studies have empirically examined the roles of key principles (e.g., fairness, accountability, transparency, explainablity) in shaping perceptions of and experience with the technologies (e.g., Shin, 2021a). Understanding the role of trust in relation to these other well-studied determinants of technology acceptance provides a useful theoretical framework for research and practice in trustworthy AI. Moreover, understanding the nature of the different dimensions underlying the concept of trust provides a fuller integration of the literature on trust with the literature on technology acceptance.

Accordingly, the primary goal of this study is to examine the role of trust as a holistic construct within the technology acceptance model (TAM) framework. In addition, we explore the core dimensions of trust perceptions and their relationship to the determinants of the acceptance of AI technologies.

The TAM is a theoretical framework for explaining the technology usage behavior that has been validated in different technologies across various populations (Venkatesh & Davis, 2000). The original TAM proposed perceived usefulness and perceived ease of use to be the two determinants of future usage intention (Davis, 1989). Subsequent studies have incorporated other external factors into the TAM, and trust has been examined as a key factor in the acceptance of emerging technologies (K. Wu et al., 2011). Trust combines characteristics, intentions, and behaviors (J. D. Lee & See, 2004; Mayer et al., 1995) and is required to build mutuality and interdependence between parties. Building on the key literature on organizational trust (Mayer et al., 1995) and trust in technology (Mcknight et al., 2011), which offer three core dimensions (benevolence/helpfulness, integrity/reliability, and competence/functionality), this paper offers a twofold conceptualization of trust: 1) trust in the human-like aspects of AI, which pertains largely to the character of the technology, and 2) trust in the functionality of AI, such as its ability, reliability and safety.

In Study 1, we examine the role of trust in the acceptance and use of AI conversational agents such as Siri, Alexa, and Google using data from a convenience sample of college students. After confirming the role of trust in the acceptance of conversational agents, in Study 2, we replicated the findings with a nationally representative sample and added the following extensions: (1) instead of measuring trust just in AI conversational agents, we examined trust in smart technologies in general, and (2) we extracted the technical and human-like dimensions of AI using exploratory factor analysis (Mayer et al., 1995; Mcknight et al., 2011) and evaluated them separately to predict intentions to use smart technologies.

By confirming the significant role of trust and demonstrating its added utility to TAM, we suggest that trust and the TAM framework can be applied to AI-driven applications, especially technologies with greater risks and/or human-like characteristics. Moreover, the multidimensional approach to trust in AI and the related measurements will be useful for researchers interested in predicting and understanding the role of human trust in AI.

**Trustworthy AI**

The uniqueness of AI is characterized by the integration of its functionality with human-like capabilities (Krafft et al., 2020). In this study, we adopt the following definition of AI offered by the OECD (2019, pp. 23–24): "An AI system is a machine-based system that can, for a given set of human-defined objectives, make predictions, recommendations, or decisions influencing real or virtual environments. AI systems are designed to operate with varying levels of autonomy."

Unlike traditional technologies in which users have complete control over their functioning, AI's capacity for autonomous functioning creates risks and uncertainties for users. Moreover, AI relies heavily on machine learning algorithms, which have been characterized as black boxes because of the inner workings of these algorithms are not easily explainable, and the way in which artificial agents learn and attain a particular answer may not be comprehensible to human actors. The black box nature of AI creates unpredictability and uncertainties, highlighting the importance of trust as users cope with the complexities and potential risks associated with AI's decision-making.

As corporations, governments, and citizens recognize the potential power and impact of AI as a sociotechnical system, great emphasis has been placed on building trustworthy AI. In recent years, government organizations (e.g., OECD, EU), tech companies (e.g., Google, Microsoft), and professional association (e.g., IEEE) have issued frameworks for the development and deployment of trustworthy AI. Their guidelines commonly recommend that trustworthy AI should be built into the system by design (Hagendorff, 2020; Jobin et al., 2019).

Many recent studies on trust in AI have focused on exploring factors that contribute to trustworthy AI (Gillath et al., 2021; Lockey et al., 2021; Rheu et al., 2020). Through a series of studies, Shin identified and empirically examined key factors that contribute to trustworthy AI systems and the role of trust in shaping people's perceptions of AI attributes (e.g., usefulness, performance, accuracy, credibility) in algorithm-driven technologies (Shin, 2020b, 2021b). Shin and Park (2019) examined the role of fairness, accountability, and transparency (FAT) in predicting users' satisfaction with algorithms and found that users with a lower level of trust considered FAT issues more skeptically. Later, Shin (2020b) extended the FAT framework to the fairness, accountability, transparency, and explainable (FATE) framework by introducing

"explainability" as an additional determinant of trustworthiness and perceptions of the usefulness of algorithmic recommendations. Shin's most recent studies have further explored the relationship between FATE, trust, user perceptions (Shin, 2021c, 2021e), and acceptance (Shin, 2021b). For example, Shin (2021e) examined the explainability of algorithms as a key predictor of FAT and tested whether explainability and FAT further predict users' trust in and perceptions of the functional attributes (i.e., accuracy, personalization, credibility) of algorithmic journalism. Overall, Shin's findings demonstrate FATE are core requirements of trustworthy AI and that trust plays a pivotal role in shaping positive perceptions of and user experience with AI technologies.

Anthropomorphism has also been commonly associated with trust in AI. Studies have found that the embodiment of an agent can increase trust because users are better able to perceive its social presence (K. Kim et al., 2018; Shin, 2021d; Waytz et al., 2014). Anthropomorphism has been associated with greater trust resilience, which prevents loss of trust (de Visser et al., 2016). However, other studies have reported weak or no effects of anthropomorphism on trust (Erebak & Turgut, 2019; Hancock et al., 2011). Glikson and Woolley (2020) identified factors that predict cognitive and emotional trust in AI. They described how tangibility and immediacy behaviors affect both cognitive and emotional trust in AI, while transparency, reliability, and task characteristics predict cognitive trust, and anthropomorphism predicts emotional trust in AI. Overall, the literature suggests that trust plays a critical role in the perceptions and acceptance of AI technologies. Past studies have examined trust in various applications of AI, including algorithmic journalism (Shin, 2020c), healthcare AI (M. K. Lee & Rich, 2021), and AI used for hiring and work evaluations (M. K. Lee, 2018). Among numerous applications and services equipped with AI technology, we focus on AI technologies in the form of consumer products.

This is because people interact most frequently with AI assistants, such as Siri and Alexa, or smart home systems that have been in the consumer market for long enough to build initial trust, thus allowing us to apply the concept. In the following section, we apply trust in AI to the TAM framework to validate whether trust can be another key determinant of AI technology acceptance.

**Trust and Use of AI Technologies: An Extended TAM**

The TAM offers a useful theoretical framework that can validate the aforementioned significance of trust in AI to predict the acceptance of AI. Originally developed by Davis (1989), the TAM has been widely employed and tested to understand the adoption of new technologies (McLean & Osei-Frimpong, 2019). The original TAM introduced perceived usefulness and perceived ease of use as the core concepts that explain individuals' adoption of technology (Davis, 1989). Later, more elaborate versions of the TAM (i.e., TAM 2, TAM 3) were developed, incorporating additional external factors, such as social norms and perceived enjoyment (Venkatesh & Bala, 2008; Venkatesh & Davis, 2000), and treating attitude as a predictor of behavior.

Following the traditional TAM framework, we first propose the following hypotheses:

**H1**: A positive attitude toward using an AI technology predicts a greater willingness to use it.

**H2:** The perceived ease of using an AI technology positively influences a) perception of its usefulness, b) trust perception, c) attitude, and d) usage intention.

**H3:** The perceived usefulness of an AI technology positively influences a) attitude toward using it and b) willingness to use it.

As trust is considered to be another significant predictor for the adoption of new technologies (Söllner et al., 2016), previous studies have investigated the connections between trust and the TAM. The role of trust was tested in the use of a new information system (Tung et al., 2008) and different online services, such as online games (J. Wu & Liu, 2007), banking (Suh & Han, 2002), social network sites (Sledgianowski & Kulviwat, 2009), and shopping (Gefen et al., 2003), among others. A meta-analysis (K. Wu et al., 2011) revealed that trust exhibited a significantly positive impact on the key TAM constructs. Recent studies continue to support the relevance of trust in TAM. For example, Shin (2020a) demonstrated that trust can positively influence continuous use of news recommendation system. In a later study, Shin (2021a) showed that trust in AI positively predict usefulness and ease of use of AI. Beldad and Hegner (2018) found that trust did not directly affect the usage intention regarding a health tracking app; rather, it influenced users' perceptions of the app's usefulness. Likewise, studies have found that trust affects usage intention indirectly rather than directly, through increasing usefulness and positive attitudes.

This study follows the hypotheses tested by other TAM-related studies that have proposed trust as an antecedent of perceived usefulness and subsequent to perceived ease of use while having a direct effect on attitude (Gefen et al., 2003; J. B. Kim, 2012; K. Wu et al., 2011). Accordingly, we propose the following hypotheses:

**H4:** Users' trust in an AI technology influences a) perceived usefulness and b) attitudes toward the technology.

**H5:** The influence of trust on behavioral intention will be mediated by a) perceived usefulness, b) attitude, and c) both perceived usefulness and attitude.

By integrating trust within TAM, we propose the following model to test the hypotheses stated above.

*[Figure 1 Near Here]*

**Multidimensional Approach to Trust in AI**

Trust is a fundamental human mechanism that is required to cope with vulnerability, uncertainty, complexity, and ambiguity, which collectively constitute a risk. Trust is defined as "a psychological state comprising the intention to accept vulnerability based upon positive expectations of the intentions or behavior of another" (Rousseau et al., 1998, p. 395). Traditionally, trust is tied to relationships between people and is required to build mutuality and interdependence between parties in human communication.

Despite the recognition of the importance of the various aspects of trust, few studies have addressed how to assess social and psychological trust in AI. Glikson and Woolley pointed out the "great variance in measures used to assess human trust in AI," which may "discourage researchers from collaborating and limit research implications to a specific discipline (p.651)." To address this shortcoming, we conceptualize and examine different dimensions of trust, including trustees' characteristics, intentions, and behaviors (J. D. Lee & See, 2004; Mayer et al., 1995). Table 1 summarizes the trust concepts and the basis of trust in different agents (people, technology, and automation) discussed in the literature.

*[Table 1 About Here]*

Mayer et al. (1995) define trust in humans as an amalgam of one's belief in another's ability, benevolence, and integrity. Ability refers to skills and competencies to successfully complete a given task. Benevolence pertains to whether the trustee has positive intentions that are not based purely on self-interest. And integrity describes the trustee's sense of morality and justice, such that the trusted party's behaviors are consistent, predictable, and honest. Trust is vital to understanding interpersonal interaction, as it contributes to perceived reliability and integrity.

Past studies have applied interpersonal trust to human technology relationships (Calhoun et al., 2019), especially when the technology has human-like characteristics (Gillath et al., 2021). Other studies have noted that human users have a distinctive view on interacting with technologies, and those works have suggested that the principles of trust in interpersonal relations cannot be directly applicable to human-to-machine trust (e.g., Madhavan & Wiegmann, 2007). McKnight et al. (2011) explained that trust in technology is qualitatively different from trust in people. The key distinction is that a human is a moral agent, whereas technology lacks volition and moral agency. The authors elaborated on three dimensions of trust in technology, functionality, reliability, and helpfulness, which they related to human trust. Functionality refers to the capability of the technology, which the authors likened to human ability. Reliability is the consistency of operation, which is like integrity, and helpfulness indicates whether the specific technology is useful to users and is like the benevolence dimension of human trust.

AI could be understood as one type of technology, but at the same time, it is involved in replacing tasks and decisions made by humans. In that sense, it is more than just a technology and McKnight's definition and the dimensions of trust in technology are not perfectly applicable to AI technologies. Unlike traditional technologies that rely on user input and the execution of rules programmed by a human, AI implies certain levels of autonomy. In one of McKnight's later studies, Lankton et al. (2015) demonstrated that for a technology with greater humanness, a trust-in-humans scale works better at predicting relevant outcomes than a trust-in-technology scale. This finding suggests that trust in technology is a dynamic concept that is still evolving and variable based on the context and the characteristics of the trusted agent. We consider that in the context of AI, both trust in people and trust in technology are relevant, since AI technologies are often depicted as being capable of human qualities, including reasoning and motivations, which can induce high expectations and initial trust (Glikson & Woolley, 2020).

Trust in automation is another relevant conceptual framework proposed by Lee and See (2004). Similar to trust in people and trust in technology, trust in automation is summarized in three categories: 1) performance, 2) process, and 3) purpose. Performance refers to the operational characteristics of automation, including its reliability and ability. Process concerns how suitable the automation is to achieve users' goals. In interpersonal relationships, this corresponds to the consistency of behaviors associated with adherence to a set of norms (Mayer et al., 1995). Purpose describes why the automation was developed and the designers' intent.

In the current study, building on previous work regarding trust, we focus on the two main dimensions of trust in AI: 1) human-like trust in AI, and 2) functionality trust in AI. The former dimension pertains to the social and cultural values of the algorithms and the values and ethics that undergird the design of the AI technology. The latter dimension pertains to the competency and expertise of the technological features. For human-like trust in AI, trust is in the agent, not in the specific actions and operations of the agent (J. D. Lee & See, 2004). People will exhibit greater trust in an AI technology whose algorithm is transparent and explainable to them (Shin, 2021b). Moreover, the perception of fairness and justice will be associated with human trust in AI. The two types of trust in AI are expected to be different in their antecedents and effects. We expected that both dimensions of trust would influence technology adoption and use (Schmidt et al., 2020), especially for technologies that are associated with risk and uncertainty. An important extension of the second study is the comparison between the two dimensions of trust in AI (the human-like dimension and the functionality dimension) and their influence on people's acceptance of AI technologies, which is proposed as a research question.

**RQ**: Is there a difference in the influence between the human-like dimension and the functionality dimension of trust in AI within the TAM framework?

## Study 1

In Study 1, we aimed to gain an initial understanding of the role of trust in the adoption and use of AI technologies. The primary goal of this study was to test H1–H5. Among various applications with AI technologies, we chose voice assistants or conversational agents, such as Apple's Siri, Amazon's Alexa, and Google Assistant, which have dramatically increased in use over the past few years (Terzopoulos & Satratzemi, 2020), warranting a closer examination of interaction with the technology and motivations underlying the adoption and use.

Voice assistants are designed to be more human-like (McLean & Osei-Frimpong, 2019), with their voice feature encouraging users to adopt the same social responses that they habitually use in their interpersonal relationships (Reeves & Nass, 1996). Today, with increasingly realistic and socially capable chatbots, research explores the similarities between how people treat other people and how they treat technologies (Gaudiello et al., 2016; Sundar et al., 2017). The anthropomorphic nature of voice assistants could make trust more critical. Therefore, in Study 1, we focused on trust in AI voice assistants and examined its significance in predicting the adoption and use.

## Method

### *Participants and Procedure*

An online survey was conducted from October to December 2020. Undergraduate students from a large Midwest university were invited to participate, and responses from 312 students were collected. Of the participants, 56% were female, and their ages ranged from 18 to

29 (median age was 21). Ethnicities represented included Caucasians (66%), Asians (21%),

African Americans (6%), and Hispanics (5%). Ninety-six percent of participants reported that they had used at least one type of AI voice assistant. Respondents received extra credit for completing the survey, which took about 10 minutes.

### Measures

The appendix summarizes the survey questionnaires and the reliability of the variables.

The constructs "perceived ease of use" and "perceived usefulness" were measured with five

items, each adapted from the original TAM scale (Davis, 1989). Trust in an AI voice assistant

was measured with four items that we created. Behavioral intention to use or continue to use an

AI voice assistant was measured with three items from the TAM 2 scale (Venkatesh & Davis,

2000).

### Analytic Approach

A path analysis was used to test the hypothesized relationships among the variables and

the model constructed in Figure 1. The analysis was conducted in R using the lavaan package

(Rosseel, 2012). Furthermore, to test the fit of the data, maximum likelihood estimation was

employed, using the following fit indices: the chi-square ($\chi$2) statistic, the comparative fit index

(CFI), the Tucker-Lewis index (TLI), the root mean square error of approximation (RMSEA),

and the standardized root mean residual (SRMR). If $\chi$2 is not significant, the CFI and TLI values are .90 or higher, and the SRMR and RMSEA values are .10 or less, then the model is considered to have a good fit. The statistical significance of the indirect effects was tested by examining bootstrapped ($k = 1,000$) 95% confidence intervals (CIs). Intercorrelations among the seven constructs were also determined using correlation analysis before the structural model was tested. The values in Table 2 indicate that the constructs are moderately correlated.

*[Table 2 Near Here]*

**Results**

It was hypothesized that AI voice assistant usage intention is influenced by four factors, namely, perceived ease of use, trust perception, perceived usefulness, and attitude (as illustrated in Figure 1). The model further indicates that perceived ease of use, trust, and perceived usefulness relate to one another, and the path model demonstrated a good fit with the data: $\chi2$ (1) = .32, $p$ = .58; RMSEA = .00 [.00, .12]; SRMR = .01; CFI = 1.00; TLI = 1.01.

*[Table 3 About Here]*

*[Figure 2 About Here]*

Regression estimates indicated that all four proposed predictors of AI voice assistant usage intention have statistically significant direct and indirect effects on the dependent variables. Perceived ease of use, perceived usefulness, and attitude were positive predictors of usage intention. Moreover, trust positively predicted perceived usefulness. Figure 2 illustrates the tested model with the standard path coefficients for the relationships among the variables.

The first set of hypotheses (H1–H3) examined the intention to use voice assistants using the traditional determinants of the TAM with the addition of trust. As hypothesized, a positive attitude toward using an AI voice assistant was associated with a greater usage intention ($\beta$ = .62, $p$ < .001; H1 is supported). In addition, perceived ease of use was positively associated with the perceived usefulness of a voice assistant ($\beta$ = .36, $p$ < .001), trust perception ($\beta$ = .39, $p$ < .001), attitude ($\beta$ = .26, $p$ < .001), and usage intention ($\beta$ = .19, $p$ < .05). Therefore, H2 was supported. The perceived usefulness of an AI voice assistant also positively influenced users' attitude toward it ($\beta$ = .42, $p$ < .001) and willingness to use it ($\beta$ = .20, $p$ < .01), thus supporting H3.

These findings yield support for the traditional TAM.

***[Table 4 Near Here]***

The second set of hypotheses (H4 and H5) examined the direct and indirect effects of trust in the TAM. The results of indirect effects revealed that all mediated paths depicted were statistically significant, including the hypothesized (H5) paths (see Table 3). Trust was associated with increased perceived usefulness, which in turn increased usage intention (H4a).

Trust also predicted positive attitudes, which in turn influenced usage intention (H4b). Moreover,

trust predicted perceived usefulness, which in turn increased positive attitudes and usage intention (H4c). Therefore, H4 was supported.

Finally, the total effects of each predictor were calculated, and perceived ease of use had the largest total effect on usage intention ($\beta$ = .63; $p$ < .001), followed by perceived usefulness ($\beta$

= .46; $p$ < .001) and trust perception ($\beta$ = .32; $p$ < .001).

**Discussion**

Study 1 presents an empirical test of the TAM for the acceptance of AI voice assistants.

The validated model presents a high explanatory power, with 52% of variance explained ($R^2$

= .52). The results of this study indicate that trust perception with the two TAM constructs—

perceived ease of use and perceived usefulness—significantly predicts attitudes toward and inclination to use voice assistants. The finding is in line with studies that have tested the impact of these constructs on the adoption of various forms of technology (e.g., Beldad & Hegner, 2018;

Choi & Ji, 2015).

Among the key predictors, perceived usefulness had a greater direct effect on usage intention than perceived ease of use. However, the total effect of perceived ease of use was greater than that of perceived usefulness. This finding suggests that to motivate individuals to adopt or continue to use an AI voice assistant, they should find the technology easy to use and useful. Thus, voice assistants should not be complicated to use, and they should provide users with useful functionalities and the benefits that people expect from using them.

In addition, we found that trust can influence and is influenced by the factors included in the TAM. Perceived ease of use contributes to trust, and trust in AI voice assistants, in turn, predicts positive attitudes and perceived usefulness, which were associated with greater usage intention. Although trust does not directly affect usage intention, the construct appears to be pivotal in improving user perception of the utility of voice assistants and in building positive attitudes toward them. An implication of this result is that people are inclined to regard a technology as beneficial if they trust it. In contrast, a lack of trust could raise concerns about the potential threats and risks of the technology instead of its benefits. Therefore, designing a trustworthy AI technology is vital, along with improving its functionality and providing an easy-to-use interface. Also, current voice assistants are designed to be more human-like than previous attempts, and many individuals are communicating with voice assistants as part of their everyday life in the same way they interact with other humans (Sundar et al., 2017), making trust more relevant.

## Study 2

In the second study, we offer an in-depth analysis of the dimensions of trust in AI within the TAM framework (the RQ). In addition, the findings from the first study (H1–H5) are replicated with a representative national sample. Study 2 extends the previous study's findings by 1) treating and measuring trust as a multidimensional construct, 2) recruiting participants that are representative of the general U.S. population, and 3) extending the acceptance of AI-based voice agents to all consumer-based AI, grouped as smart technologies.

**Method**

*Participants and Procedure*

An online survey was conducted in April 2021 using the national Qualtrics panel. To ensure the representativeness of participants, we adopted a quota sampling method that set quotas for gender, age, and race based on U.S. Census data. A total of 640 respondents (50.1% female, age $M = 46.43$, $SD = 17.83$, 65.9% White, 12% Black or African American, 12% Hispanic, 5.9% Asian, 4.4% other) participated in the study. The average time for survey completion was 10 minutes.

*Measures*

The same TAM-related items (i.e., perceived ease of use, perceived usefulness, attitude, behavior intention) from Study 1 were used in Study 2 (see the appendix). The items to measure trust in AI were different from Study 1 as we constructed the items with the consideration of three pillars of the construct used in trust in humans (Mayer et al., 1995) and trust in technology (Mcknight et al., 2011): benevolence/helpfulness, integrity/reliability, and competence/functionality. Items from the three constructs were factor analyzed, which yielded a two-factor structure, based on a principal components exploratory factor analysis (EFA) with an Oblimin rotation. The items measuring benevolence and integrity were grouped together, and competence was treated as another dimension. We named the first dimension as human-like trust in AI (six items), and the second dimension a functionality trust in AI (five items).

*[Table 5 near here]*

*Analytic Approach*

Two separate path models examined the human-like trust and functionality-trust in AI using the same analytic approach from Study 1. Intercorrelations among the constructs are presented in Table 6.

**[Table 6 Near Here]**

**Results**

H1–H5 and the RQ were addressed with two path models depicted in Figures 3 and 4. The models fit the data well: human trust in AI, $\chi^2(1) = 4.92$, $p = .03$; RMSEA = .08 [.02, .15]; SRMR = .01; CFI = 1.00; TLI = .98; technology-related trust in AI, $\chi^2(1) = 4.85$, $p = .03$ RMSEA = .08 [.02, .15]; SRMR = .01; CFI = 1.00; TLI = .98. Although the chi-square test was significant, it is highly sensitive to a large sample size. All other measures of fit (CFI above .95, SRMR below .05, and RMSEA below .08) showed that the models were adequate in reproducing covariances among the variables.

We hypothesized that people's intention to continue using smart technologies is influenced by four factors: perceived ease of use, trust (the human-like dimension and the functionality dimension), perceived usefulness, and attitude. Consistent with the findings from Study 1, regression estimates indicated that all proposed predictors of intention to use smart technologies have a statistically significant effect.

**[Figures 3 and 4 Near Here]**

**[Tables 7 and 8 About Here]**

Figure 3 and Figure 4 show the standardized path coefficients for the relationships among the variables. In both models, attitude was a statistically significant predictor of intention to use smart technologies ($\beta = .65$, $p < .001$), thus supporting H1. Furthermore, as hypothesized, perceived ease of use was positively associated with perceived usefulness ($\beta = .46$, $p < .001$; $\beta$

= .44, $p < .001$), both trust dimensions ($\beta = .56$, $p < .001$; $\beta = .51$, $p < .001$), attitude ($\beta = .16$, $p < .001$; $\beta = .17$, $p < .001$), and usage intention ($\beta = .14$, $p < .001$), thus offering robust evidence for H2. Finally, the perceived usefulness of AI smart technologies predicted attitude ($\beta = .54$, $p < .001$; $\beta = .53$, $p < .001$) and usage intention ($\beta = .22$, $p < .001$) in both models (H3 supported). These findings are in line with the initial findings from Study 1 and support the TAM's applicability to users' acceptance of AI technologies in smart objects used by consumers.

    *[Tables 9 and 10 About Here]*

    Regarding the role of the two trust dimensions in the TAM, both the human-like and the functionality dimensions of trust had statistically significant indirect effects on the intention to use AI smart technologies, thus supporting H5. The two types of trust were associated with perceived usefulness, which predicted greater usage intention (H4a). Trust also predicted positive attitudes, which in turn affected usage intention (H4b). Finally, trust dimensions predicted perceived usefulness, which in turn increased positive attitudes and usage intention (H4c). Therefore, additional support was gained for H4.

    In both path models, participants' intention to use a smart technology was predicated on its perceived ease of use (total effect $\beta = .71$; $p < .001$; $\beta = .71$; $p < .001$) and perceived usefulness (total effect $\beta = .57$; $p < .001$; $\beta = .56$; $p < .001$). In answering the RQ, there was no clearly discernable difference between the human-like trust and functionality-related trust in their pattern and magnitude of the relationships with other TAM constructs. However, the total effects of the trust dimensions revealed that the functionality-related trust dimension exhibited a greater total impact on usage intention than the human-like trust dimension (total effect of human-like trust: $\beta = .36$; $p < .001$; total effect of functionality-related trust: $\beta = .42$; $p < .001$).

**Discussion**

The results of the second study with the general population provide additional support for the key determinants (perceived ease of use, usefulness, and trust) of a user's attitude and behavioral intention associated with the acceptance of AI technologies suggested by the extended TAM. These findings are consistent with Study 1 and prior literature (Gefen et al., 2003; I.-L. Wu & Chen, 2005).

Additionally, the results indicate that the two dimensions of trust in AI—the *human-like trust dimension and the functionality dimension*—significantly predict perceived usefulness and positive attitude toward smart technology, which in turn, predict greater usage intention. Both human-like trust in AI and functionality-related trust in AI had a positive impact on perceived usefulness, attitude, and usage intention, and the functionality dimension of trust had a greater total impact than the human-like trust dimension. This suggests that while both trust constructs can be useful in understanding the multidimensional aspects and measurement of trust in AI, trust in the functionality of AI is particularly useful when applying the TAM to understand AI acceptance. At the same time, the significance of the human-like properties of AI, such as social graces, protection of privacy, protocols for ensuring fairness and avoiding bias were also significant. It appears the duality of AI as both a functional and social technology must be considered when designing trustworthy AI. Further, human-like trust may explain emotional trust and emotional attachment to technology.

The distinctive perceptions of human-like trust and functionality trust align with studies that propose multiple dimensions and factors of trust in AI and automation (Glikson & Woolley, 2020; Hoff & Bashir, 2015; J. D. Lee & See, 2004). For instance, Glikson and Woolley (2020) organized the literature on trust in AI based on the factors that shaped users' cognitive and emotional trust. The cognitive construct involves the rational evaluation of AI while emotional trust is typically influenced by irrational factors.

**General Discussion**

Our findings provide further support for the utility of the TAM in explaining the acceptance of AI technologies and the added contribution of trust within the model. The addition of trust to the TAM model was examined in two empirical studies conducted with different AI technology applications and respondent populations. In Study 1 and Study 2, we found that perceived ease of use and perceived usefulness were essential determinants for the use of AI technologies, although perceived ease of use had a consistently greater impact on the acceptance of these technologies. This impact may be because consumers tend to believe that AI technologies are complicated and difficult to use. The model explains that when people consider an AI technology to be easy to use, they are more likely to trust the technology and consider it useful. Therefore, for AI-based consumer technologies, making the technology easy and straightforward enough to operate without much difficulty is critical for its uptake.

Our studies confirmed the significant role of trust in shaping people's attitudes and acceptance of AI technologies with the TAM framework. As past research has already confirmed, trust in AI can be built through embedding specific principles in the design and governance of AI, as suggested in recent publications using the fairness, accountability, transparency, explainability framework (Shin, 2020b; Shin & Park, 2019). In addition to the key attributes of trustworthy AI, our findings highlight the importance of the experiential aspect of AI technology to creating and maintaining trust. Indeed, trust is relevant throughout the lifecycle of a technology. Users who participate in an experience with technology build ideas about whether it is useful and easy to use based on their experience of use. Thus, the concept of trustworthy AI should not be limited to the design but should also integrate the direct experience of users, and the TAM offers a useful framework in that regard.

Overall, our findings suggest that future studies investigating the acceptance of AI must include the trust construct as an integral component of their predictive models. Given the central role of trust, there is a strong practical need to understand what facilitates trust in AI and what contributes to the design of trustworthy AI.

Additionally, this study offers concepts and measurements for two dimensions of trust in AI. We operationalized trust in AI with a human-like dimension to capture the automatic priming of social cues, intentionality, and moral agency, although AI not human. This dimension was a significant predictor of the acceptance of technology, as was the functionality dimension. The distinction between human-like trust and functionality-related trust in AI, and their measures tested in this study, will be useful for future AI research, as trust will be even more critical in AI applications in high-stake domains such as autonomous vehicles and medical decision-making.

We also urge researchers to reflect upon the various dimensions of trust because our study findings show a clear perceptual difference exists between the two dimensions. Hence, studies that improve our understanding of the antecedents and effects of human-like and technology-related trust in AI will be useful. Another useful approach, not covered in this study, is that of multi-level or multi-layer trust in AI. Hoff and Bashir (2015) suggested three layers of trust in automation (dispositional trust, situational trust, and learned trust) based on their review of trust in AI. Such trust models offer additional insights into research on trustworthy AI.

## Conclusion

Given the progress that has been made in AI development, a more sophisticated understanding of trust in AI is required than was previously the case. This study examined different dimensions of trust within the established TAM framework, which has been used widely to examine the acceptance of various new technologies (Marangunić & Granić, 2015). In the days ahead, the influence of AI on everyday life is expected to continuously grow, and given the black box nature of the underlying mechanisms of AI, the importance of trust in AI is also likely to grow. The key contributions of our study are summarized below.

**Theoretical Contributions**

Our core theoretical contribution lies in extending the classical TAM framework by including trust-specific aspects of AI. Overall, our findings suggest that future studies investigating the acceptance of AI must include the trust construct as an integral part of their predictive models. Given the central role of trust, there is a strong need to understand what facilitates trust in AI and what contributes to the design of trustworthy AI.

In addition, the TAM was used to assess the effects of different dimensions of trust. The multidimensional approach provides a new lens for conceptualizing the variability of people's trust perceptions in AI. Its structure can be applied to help guide future research. For example, specific ethical principles of AI may align with specific dimensions of trust. The trust-building factors already outlined in previous research may be closely related to a specific dimension of trust. For example, anthropomorphism may increase human-like trust perceptions, yet it may have no impact on perceptions of functionality. Similarly, people may not simultaneously possess either high or low trust in both dimensions of trust. In such cases, understanding the effects of a specific type of trust can be useful. Therefore, we urge future research to examine trust in AI by including its different dimensions.

**Practical Contributions**

This research also contributes to the growing body of practice that seeks to understand the role of trust in accepting AI technologies. Our findings suggest that AI technologies should be easy to use, useful, and trusted. When such characteristics are coupled with other factors that contribute to trustworthy perceptions of AI (e.g., FATE, anthropomorphism), the technologies are likely to be accepted and used by many people. As much as the designers of AI systems care about making the systems useful and easy to use, designing AI embedded with values that makes them trustworthy should be a critical design concern.

By conceptualizing and examining possible dimensions of trust in AI, this study contributes to understanding how to ensure such abstract trust issues in AI are addressed; how to design AI systems that are value-oriented, ethical, and human-centered; and how to govern AI in support of trustworthy AI systems. Moreover, our multidimensional framework can be used to develop interventions and design procedures that encourage appropriate levels of trust.

**Limitations and Future Research Directions**

This study has some limitations that offer promising research directions for the future. First, we focused on AI technology applications in the form of consumer products (i.e., voice assistants and smart technologies), which tend be associated with lower risks. Future research could further explore high-stakes AI (e.g., self-driving cars, AI in healthcare) and different AI-powered applications with other human-like characteristics, such as robots and embedded algorithms, to examine the role of trust and additional critical factors in the adoption and use of such technologies.

Second, trust is a dynamic concept that varies depending on the stage of technology. For example, the trust life cycle can be divided into initial trust and ongoing trust, depending on one's initial experience, and these forms of trust may play different roles (J. B. Kim, 2012). Future studies could also develop measurement scales for differentiated targets of trust (e.g., technology itself, the company, or governing bodies; Söllner et al., 2016).

      Third, our path models are based on correlations and cannot guarantee causal effects of trust on the acceptance of AI technologies. We proposed trust as an antecedent of perceived usefulness based on the findings from the past TAM literature (K. Wu et al., 2011). Still, an opposite direction of the relationship was also found to be valid in another study (Shin, 2020a). Therefore, future studies can test a causal model to determine the exact relationships among TAM constructs to extend current findings.

      Finally, as with any online panel, our study respondents may not be truly representative of the general population. The fact that all the study participants had access to computers and the internet skewed the sample toward a higher acceptance of technology. Participants in an online survey are likely to have greater access to and experience with various forms of information technology. Therefore, future studies should examine more representative samples with varying levels of experience with technologies.

**References**

AI HLEG. (2019). *Ethics guidelines for trustworthy AI*. European Commission.

  https://data.europa.eu/doi/10.2759/346720

Barredo Arrieta, A., Díaz-Rodríguez, N., Del Ser, J., Bennetot, A., Tabik, S., Barbado, A.,

  Garcia, S., Gil-Lopez, S., Molina, D., Benjamins, R., Chatila, R., & Herrera, F. (2020).

  Explainable Artificial Intelligence (XAI): Concepts, taxonomies, opportunities and

  challenges toward responsible AI. *Information Fusion*, *58*, 82–115.

  https://doi.org/10.1016/j.inffus.2019.12.012

Beldad, A. D., & Hegner, S. M. (2018). Expanding the Technology Acceptance Model with the

  Inclusion of Trust, Social Influence, and Health Valuation to Determine the Predictors of

  German Users' Willingness to Continue using a Fitness App: A Structural Equation

  Modeling Approach. *International Journal of Human–Computer Interaction*, *34*(9), 882–

  893. https://doi.org/10.1080/10447318.2017.1403220

Calhoun, C. S., Bobko, P., Gallimore, J. J., & Lyons, J. B. (2019). Linking precursors of

  interpersonal trust to human-automation trust: An expanded typology and exploratory

  experiment. *Journal of Trust Research*, *9*(1), 28–46.

  https://doi.org/10.1080/21515581.2019.1579730

Choi, J. K., & Ji, Y. G. (2015). Investigating the Importance of Trust on Adopting an

  Autonomous Vehicle. *International Journal of Human-Computer Interaction*, *31*(10),

  692–702. https://doi.org/10.1080/10447318.2015.1070549

Davis, F. D. (1989). Perceived Usefulness, Perceived Ease of Use, and User Acceptance of

  Information Technology. *MIS Quarterly*, *13*(3), 319. https://doi.org/10.2307/249008

de Visser, E. J., Monfort, S. S., McKendrick, R., Smith, M. A. B., McKnight, P. E., Krueger, F.,

    & Parasuraman, R. (2016). Almost human: Anthropomorphism increases trust resilience

    in cognitive agents. *Journal of Experimental Psychology: Applied*, *22*(3), 331–349.

    https://doi.org/10.1037/xap0000092

Edelman. (2021). *Edelman trust barometer 2021* (Annual Edelman Trust Barometer, p. 58).

    https://www.edelman.com/sites/g/files/aatuss191/files/2021-

    03/2021%20Edelman%20Trust%20Barometer.pdf

Erebak, S., & Turgut, T. (2019). Caregivers' attitudes toward potential robot coworkers in elder

    care. *Cognition, Technology & Work*, *21*(2), 327–336. https://doi.org/10.1007/s10111-

    018-0512-0

Gaudiello, I., Zibetti, E., Lefort, S., Chetouani, M., & Ivaldi, S. (2016). Trust as indicator of

    robot functional and social acceptance. An experimental study on user conformation to

    iCub answers. *Computers in Human Behavior*, *61*, 633–655.

    https://doi.org/10.1016/j.chb.2016.03.057

Gefen, D., Karahanna, E., & Straub, D. W. (2003). Trust and TAM in Online Shopping: An

    Integrated Model. *MIS Quarterly*, *27*(1), 51–90. https://doi.org/10.2307/30036519

Gillath, O., Ai, T., Branicky, M. S., Keshmiri, S., Davison, R. B., & Spaulding, R. (2021).

    Attachment and trust in artificial intelligence. *Computers in Human Behavior*, 10.

Glikson, E., & Woolley, A. W. (2020). Human trust in artificial intelligence: Review of

    empirical research. *Academy of Management Annals*, *14*(2), 627–660.

    https://doi.org/10.5465/annals.2018.0057

Gorwa, R., Binns, R., & Katzenbach, C. (2020). Algorithmic content moderation: Technical and political challenges in the automation of platform governance. *Big Data & Society*, *7*(1),

2053951719897945. https://doi.org/10.1177/2053951719897945

Hagendorff, T. (2020). The ethics of AI ethics: An evaluation of guidelines. *Minds and

Machines*, *30*(1), 99–120. https://doi.org/10.1007/s11023-020-09517-8

Hancock, P. A., Billings, D. R., Schaefer, K. E., Chen, J. Y. C., de Visser, E. J., & Parasuraman,

R. (2011). A Meta-Analysis of Factors Affecting Trust in Human-Robot Interaction.

*Human Factors*, *53*(5), 517–527. https://doi.org/10.1177/0018720811417254

Hoff, K. A., & Bashir, M. (2015). Trust in automation: Integrating empirical evidence on factors

that influence trust. *Human Factors*, *57*(3), 407–434.

https://doi.org/10.1177/0018720814547570

Jobin, A., Ienca, M., & Vayena, E. (2019). The global landscape of AI ethics guidelines. *Nature

Machine Intelligence*, *1*(9), 389–399. https://doi.org/10.1038/s42256-019-0088-2

Kim, J. B. (2012). An empirical study on consumer first purchase intention in online shopping:

Integrating initial trust and TAM. *Electronic Commerce Research*, *12*(2), 125–150.

https://doi.org/10.1007/s10660-012-9089-5

Kim, K., Boelling, L., Haesler, S., Bailenson, J., Bruder, G., & Welch, G. F. (2018). Does a

digital assistant need a body? The influence of visual embodiment and social behavior on

the perception of intelligent virtual agents in AR. *2018 IEEE International Symposium on

Mixed and Augmented Reality (ISMAR)*, 105–114.

https://doi.org/10.1109/ISMAR.2018.00039

Krafft, P. M., Young, M., Katell, M., Huang, K., & Bugingo, G. (2020). Defining AI in Policy versus Practice. *Proceedings of the AAAI/ACM Conference on AI, Ethics, and Society*, 72–78. https://doi.org/10.1145/3375627.3375835

Lankton, N., McKnight, D. H., & Tripp, J. (2015). Technology, humanness, and trust: Rethinking trust in technology. *Journal of the Association for Information Systems*, *16*(10), 880–918. https://doi.org/10.17705/1jais.00411

Lee, J. D., & See, K. A. (2004). Trust in automation: Designing for appropriate reliance. *Human Factors*, *46*(1), 50–80. https://doi.org/10.1518/hfes.46.1.50_30392

Lee, M. K. (2018). Understanding perception of algorithmic decisions: Fairness, trust, and emotion in response to algorithmic management. *Big Data & Society*, *5*(1), 2053951718756684. https://doi.org/10.1177/2053951718756684

Lee, M. K., & Rich, K. (2021). *Who is included in human perceptions of AI?: Trust and perceived fairness around healthcare AI and cultural mistrust*. 14.

Lockey, S., Gillespie, N., Holm, D., & Someh, I. A. (2021). *A review of trust in artificial intelligence: Challenges, vulnerabilities and future directions*. Hawaii International Conference on System Sciences. https://doi.org/10.24251/HICSS.2021.664

Madhavan, P., & Wiegmann, D. A. (2007). Effects of information source, pedigree, and reliability on operator interaction with decision support systems. *Human Factors: The Journal of the Human Factors and Ergonomics Society*, *49*(5), 773–785. https://doi.org/10.1518/001872007X230154

Marangunić, N., & Granić, A. (2015). Technology acceptance model: A literature review from 1986 to 2013. *Universal Access in the Information Society*, *14*(1), 81–95. https://doi.org/10.1007/s10209-014-0348-1

Mayer, R. C., Davis, J. H., & Schoorman, F. D. (1995). An integrative model of organizational trust: Past, present, and future. *The Academy of Management Review*, *20*(3), 709–734.

Mcknight, D. H., Carter, M., Thatcher, J. B., & Clay, P. F. (2011). Trust in a specific technology: An investigation of its components and measures. *ACM Transactions on Management Information Systems*, *2*(2), 1–25. https://doi.org/10.1145/1985347.1985353

McLean, G., & Osei-Frimpong, K. (2019). Hey Alexa … examine the variables influencing the use of artificial intelligent in-home voice assistants. *Computers in Human Behavior*, *99*, 28–37. https://doi.org/10.1016/j.chb.2019.05.009

OECD. (2019). *Artificial intelligence in society*. OECD Publishing. https://doi.org/10.1787/eedfee77-en

Reeves, B., & Nass, C. I. (1996). *The media equation: How people treat computers, television, and new media like real people and places* (1. paperback ed). CSLI Publ.

Rheu, M., Shin, J. Y., Peng, W., & Huh-Yoo, J. (2020). Systematic Review: Trust-Building Factors and Implications for Conversational Agent Design. *International Journal of Human–Computer Interaction*, 1–16. https://doi.org/10.1080/10447318.2020.1807710

Rosseel, Y. (2012). **lavaan**: An *R* Package for Structural Equation Modeling. *Journal of Statistical Software*, *48*(2). https://doi.org/10.18637/jss.v048.i02

Rousseau, D. M., Sitkin, S. B., Burt, R. S., & Camerer, C. (1998). Not so different after all: A cross-discipline view of trust. *Academy of Management Review*, *23*(3), 393–404. https://doi.org/10.5465/amr.1998.926617

Russell, S. J., Norvig, P., Davis, E., & Edwards, D. (2016). *Artificial intelligence: A modern approach* (Third edition, Global edition). Pearson.

Schmidt, P., Biessmann, F., & Teubner, T. (2020). Transparency and trust in artificial

    intelligence systems. *Journal of Decision Systems*, *29*(4), 260–278.

    https://doi.org/10.1080/12460125.2020.1819094

Shin, D. (2020a). How do users interact with algorithm recommender systems? The interaction

    of users, algorithms, and performance. *Computers in Human Behavior*, *109*, 106344.

    https://doi.org/10.1016/j.chb.2020.106344

Shin, D. (2020b). User perceptions of algorithmic decisions in the personalized AI

    system:Perceptual evaluation of fairness, accountability, transparency, and explainability.

    *Journal of Broadcasting & Electronic Media*, *64*(4), 541–565.

    https://doi.org/10.1080/08838151.2020.1843357

Shin, D. (2020c). Expanding the Role of Trust in the Experience of Algorithmic Journalism:

    User Sensemaking of Algorithmic Heuristics in Korean Users. *Journalism Practice*, 1–

    24. https://doi.org/10.1080/17512786.2020.1841018

Shin, D. (2021a). Embodying algorithms, enactive artificial intelligence and the extended

    cognition: You can see as much as you know about algorithm. *Journal of Information

    Science*, 0165551520985495. https://doi.org/10.1177/0165551520985495

Shin, D. (2021b). The effects of explainability and causability on perception, trust, and

    acceptance: Implications for explainable AI. *International Journal of Human-Computer

    Studies*, *146*, 102551. https://doi.org/10.1016/j.ijhcs.2020.102551

Shin, D. (2021c). How do people judge the credibility of algorithmic sources? *AI & SOCIETY*.

    https://doi.org/10.1007/s00146-021-01158-4

Shin, D. (2021d). The perception of humanness in conversational journalism: An algorithmic

    information-processing perspective. *New Media & Society*, 1461444821993801.

    https://doi.org/10.1177/1461444821993801

Shin, D. (2021e). Why does explainability matter in news analytic systems? Proposing

    explainable analytic journalism. *Journalism Studies*, *22*(8), 1047–1065.

    https://doi.org/10.1080/1461670X.2021.1916984

Shin, D., & Park, Y. J. (2019). Role of fairness, accountability, and transparency in algorithmic

    affordance. *Computers in Human Behavior*, *98*, 277–284.

    https://doi.org/10.1016/j.chb.2019.04.019

Söllner, M., Hoffmann, A., & Leimeister, J. M. (2016). Why different trust relationships matter

    for information systems users. *European Journal of Information Systems*, *25*(3), 274–

    287. https://doi.org/10.1057/ejis.2015.17

Sundar, S. S., Jung, E. H., Waddell, T. F., & Kim, K. J. (2017). Cheery companions or serious

    assistants? Role and demeanor congruity as predictors of robot attraction and use

    intentions among senior citizens. *International Journal of Human-Computer Studies*, *97*,

    88–97. https://doi.org/10.1016/j.ijhcs.2016.08.006

Terzopoulos, G., & Satratzemi, M. (2020). Voice Assistants and Smart Speakers in Everyday

    Life and in Education. *Informatics in Education*, 473–490.

    https://doi.org/10.15388/infedu.2020.21

Thiebes, S., Lins, S., & Sunyaev, A. (2020). Trustworthy artificial intelligence. *Electronic

    Markets*, 18. https://doi.org/10.1007/s12525-020-00441-4

Turing, A. M. (1950). Computing machinery and intelligence. *Mind*, *LIX*(236), 433–460.

    https://doi.org/10.1093/mind/LIX.236.433

Venkatesh, V., & Davis, F. D. (2000). A theoretical extension of the Technology Acceptance

    Model: Four longitudinal field studies. *Management Science*, *46*(2), 186–204.

    https://doi.org/10.1287/mnsc.46.2.186.11926

Watson, D. (2019). The rhetoric and reality of anthropomorphism in artificial intelligence. *Minds*

    *and Machines*, *29*(3), 417–440. https://doi.org/10.1007/s11023-019-09506-6

Waytz, A., Heafner, J., & Epley, N. (2014). The mind in the machine: Anthropomorphism

    increases trust in an autonomous vehicle. *Journal of Experimental Social Psychology*, *52*,

    113–117. https://doi.org/10.1016/j.jesp.2014.01.005

Wu, I.-L., & Chen, J.-L. (2005). An extension of Trust and TAM model with TPB in the initial

    adoption of on-line tax: An empirical study. *International Journal of Human-Computer*

    *Studies*, *62*(6), 784–808. https://doi.org/10.1016/j.ijhcs.2005.03.003

Wu, K., Zhao, Y., Zhu, Q., Tan, X., & Zheng, H. (2011). A meta-analysis of the impact of trust

    on technology acceptance model: Investigation of moderating influence of subject and

    context type. *International Journal of Information Management*, *31*(6), 572–581.

    https://doi.org/10.1016/j.ijinfomgt.2011.03.004

**Tables and Figures**

**Table 1.**

*Dimensions of Trust*

| Basis of trust in people<br>Mayer et al., (1995) | Basis of trust in technology<br>Lankton et al. (2015) | Basis of trust in automation<br>Lee & Moray (1992) |
|---|---|---|
| Competence/Ability | Functionality | Performance |
| Integrity | Reliability | Process |
| Benevolence | Helpfulness | Purpose |

**Table 2.**

*Zero-Order Correlations, Means, and Standard Deviations (Study 1)*

| Variable | 1 | 2 | 3 | 4 | 5 |
|---|---|---|---|---|---|
| 1. Ease of use | 1 | | | | |
| 2. Trust | 38** | 1 | | | |
| 3. Usefulness | .46** | .48** | 1 | | |
| 4. Attitude | .56** | .56** | .70** | 1 | |
| 5. Behavior Intention | .50** | .40** | .59** | .70** | 1 |
| *M* | 3.81 | 3.03 | 3.42 | 3.57 | 3.77 |
| *SD* | .81 | .83 | .91 | .82 | 1.02 |

*Note.* ** $p < .01$

**Table 3.**

*Standardized Regression Path Analysis (Study 1)*

| Paths | $\beta$ | *p*-value | Hypotheses |
|---|---|---|---|
| PE → PU | .36 | <.001 | H2a |
| PE → TR | .39 | <.001 | H2b |
| PE → ATT | .26 | <.001 | H2c |
| PE → BI | .19 | .012 | H2d |
| PU → ATT | .42 | <.001 | H3a |
| PU → BI | .20 | .007 | H3b |
| TR → PU | .39 | <.001 | H4a |
| TR → ATT | .23 | <.001 | H4b |
| ATT → BI | .62 | <.001 | H1 |

*Note.* PE = Perceived ease of use; PU = Perceived usefulness; TR = Trust; ATT = Attitude; BI = Behavior intention

**Table 4.**

*Analysis of Indirect Effects (Study 1)*

| Paths | Effect (95% CI) | LLCI | ULCI | Hypotheses |
|---|---|---|---|---|
| PE → ATT → BI | .16 | .10 | .23 | |
| PE → PU → BI | .07 | .02 | .15 | |
| PE → PU → ATT → BI | .09 | .06 | .16 | |
| PE → TRU → ATT → BI | .05 | .03 | .09 | |
| PE → TRU → PU → BI | .03 | .01 | .06 | |
| PE → TRU → PU → ATT → BI | .04 | .02 | .07 | |
| PU → ATT → BI | .26 | .18 | .36 | |
| TRU → PU → BI | .08 | .02 | .15 | H4a |
| TRU → ATT → BI | .14 | .08 | .22 | H4b |
| TRU → PU→ ATT → BI | .03 | .06 | .16 | H4c |

*Note.* PE = Perceived ease of use; PU = Perceived usefulness; TR = Trust; ATT = Attitude; BI = Behavior intention

**Table 5.**

*Results From an Exploratory Factor Analysis of the Trust in AI Questionnaire (Study 2)*

| Trust in AI items | Factor loading | | Communality |
|---|---|---|---|
| | 1 | 2 | |
| **Factor 1: Human-like trust in AI** | | | |
| Smart technologies care about our well-being. (Benevolence) | **.92** | -.08 | .75 |
| Smart technologies are sincerely concerned about addressing the problems of human users. (Benevolence) | **.89** | -.04 | .74 |
| Smart technologies try to be helpful and do not operate out of selfish interest. (Benevolence) | **.81** | -.02 | .64 |
| Smart technologies are truthful in their dealings. (Integrity) | **.79** | .09 | .73 |
| Smart technologies keep their commitments and deliver on their promises. (Integrity) | **.60** | .29 | .70 |
| Smart technologies are honest and do not abuse the information and advantage they have over their users. (Integrity) | **.82** | .05 | .73 |
| **Factor 2: Functionality trust in AI** | | | |
| Smart technologies work well. (Competence) | .04 | **.81** | .71 |
| Smart technologies have the features necessary to complete key tasks. (Competence) | .08 | **.76** | .67 |
| Smart technologies have the features necessary to complete key tasks. (Competence) | -.03 | **.87** | .72 |
| Smart technologies are reliable. (Competence) | -.06 | **.92** | .78 |
| Smart technologies are dependable. (Competence) | .03 | **.84** | .74 |

*Note.* $N = 640$. The extraction method was principal axis factoring with an oblique (promax with Kaiser normalization) rotation. Factor loadings above .60 are in bold.

**Table 6.**

*Zero-Order Correlations, Means, and Standard Deviations (Study 2)*

| Variable | 1 | 2 | 3 | 4 | 5 | 6 |
|---|---|---|---|---|---|---|
| 1. Ease of use | 1 | | | | | |
| 2. Human trust in AI | .51** | 1 | | | | |
| 3. Technology trust in AI | .54** | .75** | 1 | | | |
| 4. Usefulness | .67** | .60** | .64** | 1 | | |
| 5. Attitude | .65** | .68** | .69** | .77** | 1 | |
| 6. Behavior Intention | .66** | .57** | .65** | .76** | .85** | 1 |
| *M* | 3.65 | 3.27 | 3.64 | 3.74 | 3.69 | 3.76 |
| *SD* | .98 | 1.03 | .92 | .91 | 1.00 | 1.05 |

*Note.* ** $p < .01$

**Table 7.**

*Standardized Regression Path Analysis with Human-like Trust in AI (Study 2)*

| Paths | $\beta$ | *p*-value | Hypotheses |
|---|---|---|---|
| PE → PU | .46 | <.001 | H2a |
| PE → HTR | .56 | <.001 | H2b |
| PE → ATT | .16 | <.001 | H2c |
| PE → BI | .14 | <.001 | H2d |
| PU → ATT | .54 | <.001 | H3a |
| PU → BI | .22 | <.001 | H3b |
| HTR → PU | .30 | <.001 | H4a |
| HTR → ATT | .30 | <.001 | H4b |
| ATT → BI | .65 | <.001 | H1 |

*Note.* PE = Perceived ease of use; PU = Perceived usefulness; HTR = Human trust in AI; ATT = Attitude; BI = Behavior intention

**Table 8.**

*Standardized Regression Path Analysis* with *Functionality Trust in AI (Study 2)*

| Paths | $\beta$ | $p$-value | Hypotheses |
|-------|---------|-----------|------------|
| PE → PU | .44 | <.001 | H2a |
| PE → TTR | .51 | <.001 | H2b |
| PE → ATT | .17 | <.001 | H2c |
| PE → BI | .14 | <.001 | H2d |
| PU → ATT | .53 | <.001 | H3a |
| PU → BI | .22 | <.001 | H3b |
| TTR → PU | .38 | <.001 | H4a |
| TTR → ATT | .31 | <.001 | H4b |
| ATT → BI | .65 | <.001 | H1 |

*Note.* PE = Perceived ease of use; PU = Perceived usefulness; TTR = Technology trust in AI; ATT = Attitude; BI = Behavior intention

**Table 9.**

*Analysis of Indirect Effects with Human Trust in AI (Study 2)*

| Paths | Effect (95% CI) | LLCI | ULCI | Hypotheses |
|-------|-----------------|------|------|------------|
| PE → ATT → BI | .10 | .05 | .16 | |
| PE → PU → BI | .10 | .05 | .16 | |
| PE → PU → ATT → BI | .16 | .12 | .21 | |
| PE → HTR → ATT → BI | .11 | .08 | .15 | |
| PE → HTR → PU → BI | .04 | .02 | .06 | |
| PE → HTR → PU → ATT → BI | .06 | .04 | .08 | |
| PU → ATT → BI | .35 | .27 | .43 | |
| HTR → PU → BI | .07 | .03 | .10 | H4a |
| HTR → ATT → BI | .19 | .14 | .26 | H4b |
| HTR → PU→ ATT → BI | .10 | .07 | .14 | H4c |

*Note.* PE = Perceived ease of use; PU = Perceived usefulness; HTR = Human trust in AI; ATT = Attitude; BI = Behavior intention

**Table 10.**

*Analysis of Indirect Effects with Technology Trust in AI (Study 2)*

| Paths | Effect (95% CI) | LLCI | ULCI | Hypotheses |
|-------|-----------------|------|------|------------|
| PE → ATT → BI | .11 | .06 | .17 | |
| PE → PU → BI | .10 | .05 | .15 | |
| PE → PU → ATT → BI | .15 | .11 | .21 | |
| PE → TTR → ATT → BI | .10 | .06 | .15 | |
| PE → TTR → PU → BI | .04 | .02 | .07 | |
| PE → TTR → PU → ATT → BI | .07 | .05 | .10 | |
| PU → ATT → BI | .34 | .28 | .43 | |
| TTR → PU → BI | .08 | .04 | .14 | H4a |
| TTR → ATT → BI | .20 | .13 | .29 | H4b |
| TTR → PU→ ATT → BI | .13 | .09 | .18 | H4c |

*Note.* PE = Perceived ease of use; PU = Perceived usefulness; TTR = Technology trust in AI; ATT = Attitude; BI = Behavior intention

**Figure 1.**

*Hypothesized Path Model*

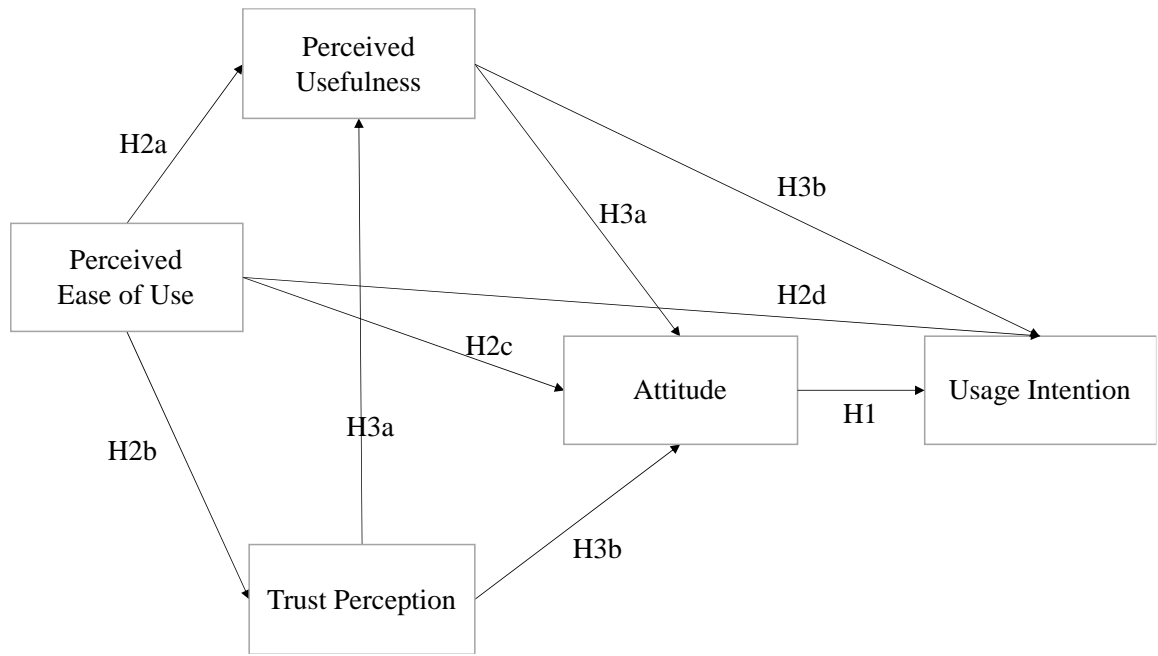

**Figure 2.**

*Results of Path Analysis (Study 1)*

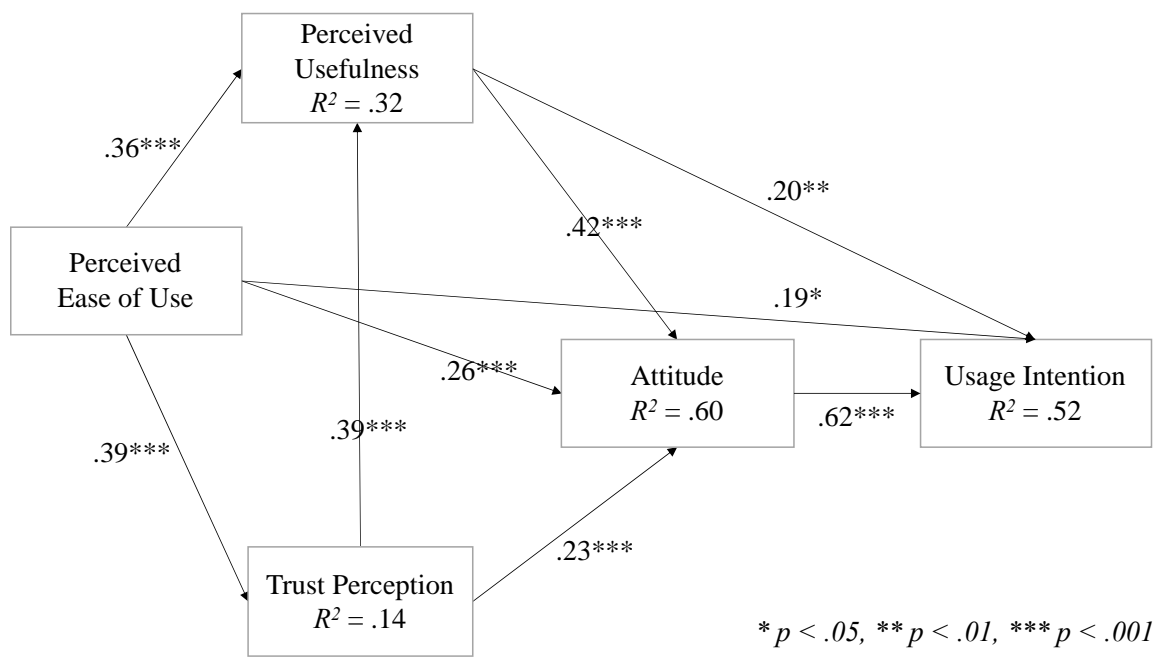

**Figure 3.**

*Results of Path Analysis with Human-like Trust in AI (Study 2)*

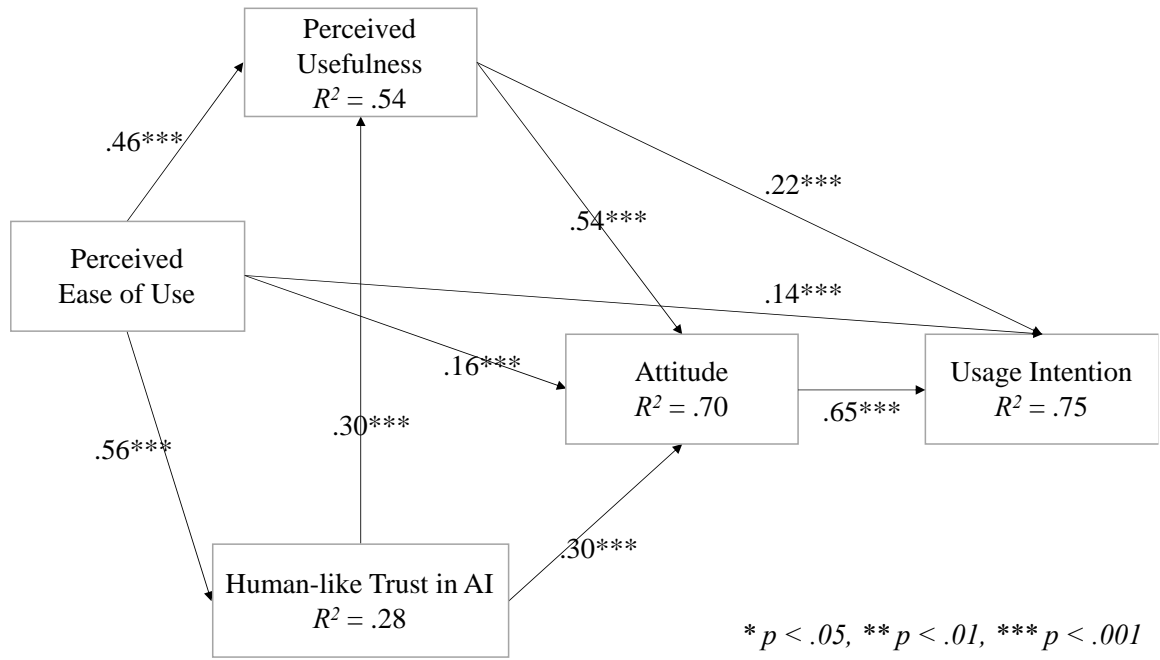

**Figure 4.**

*Results of Path Analysis with Functionality Trust in AI (Study 2)*

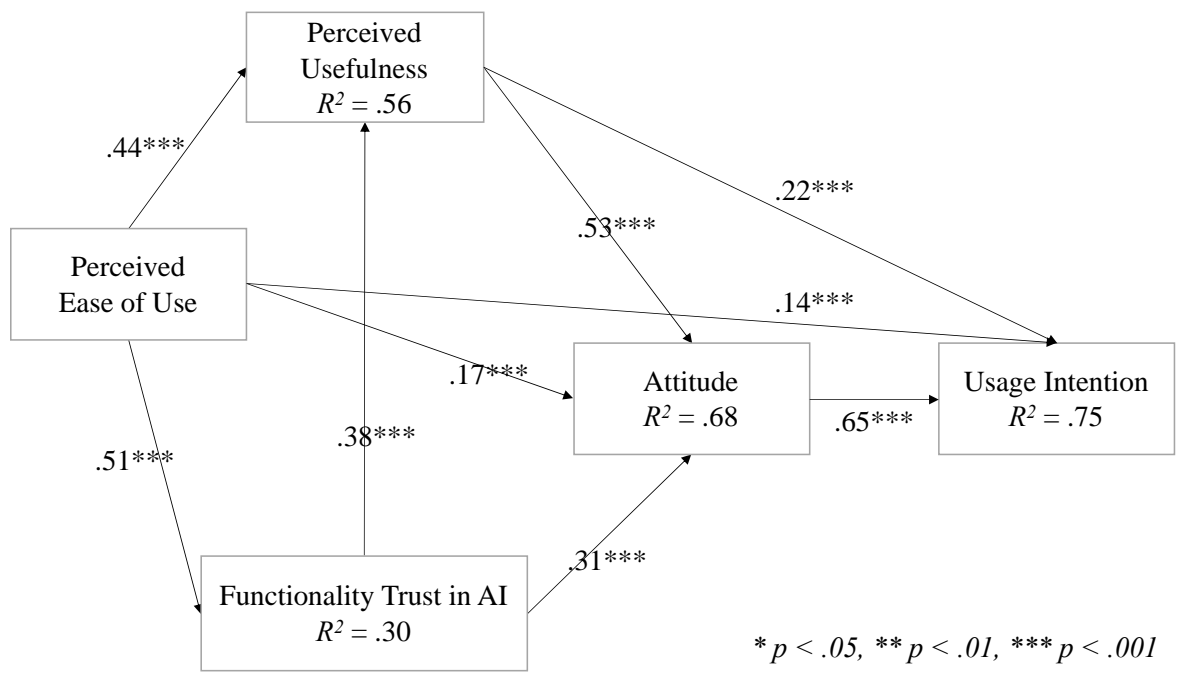

**Appendix**

*Survey Questionnaires, Scales, and Reliability Coefficients*

| Variable | Survey items | Scale | Reliability | |
|---|---|---|---|---|
| | | | Study 1 | Study 2 |
| Perceived ease of use (Study 1 & Study 2) | Learning to use [AI virtual assistants/AI smart technologies] would be easy for me<br>I would find it easy to get [AI virtual assistants/AI smart technologies] to do what I want it to do<br>My interaction with [AI virtual assistants/AI smart technologies] is clear and understandable<br>It would be easy for me to become skillful at using [AI virtual assistants/AI smart technologies]<br>I would find [AI virtual assistants/AI smart technologies] to be easy to use | 1 (strongly disagree) – 5 (strongly agree) | $\alpha = .90$ | $\alpha = .91$ |
| Trust in the voice assistant (Study 1) | I trust that AI virtual assistants can offer information and service that's best of my interest<br>I trust that my personal data is protected from potential abuse when using AI virtual assistants<br>I trust that my privacy is protected when using AI virtual assistants<br>I trust that authorities exerts effective control over organizations and companies providing AI virtual assistant services | 1 (strongly disagree) – 5 (strongly agree) | $\alpha = .81$ | N/A |
| Perceived usefulness (Study 1 & Study 2) | Using [AI virtual assistants/AI smart technologies] would enable me to accomplish tasks more quickly<br>Using [AI virtual assistants/AI smart technologies] would improve my performance at accomplishing tasks<br>Using [AI virtual assistants/AI smart technologies] for accomplishing tasks would increase my productivity<br>Using [AI virtual assistants/AI smart technologies] would enhance my effectiveness at accomplishing tasks<br>I find [AI virtual assistants/AI smart technologies] useful for me to accomplish tasks | 1 (strongly disagree) – 5 (strongly agree) | $\alpha = .92$ | $\alpha = .90$ |
| Attitude toward the AI technologies (Study 1 & Study 2) | I feel positive toward [AI virtual assistants/AI smart technologies]<br>I feel that using [AI virtual assistants/AI smart technologies] is pleasant<br>Using [AI virtual assistants/AI smart technologies] is a good idea<br>Using [AI virtual assistants/AI smart technologies] is a smart way to get things done | 1 (strongly disagree) – 5 (strongly agree) | $\alpha = .89$ | $\alpha = .90$ |
| Behavioral intention of non-users (Study 1 & Study 2) | I intend to use [AI virtual assistants/AI smart technologies] in a future<br>I predict that I would use [AI virtual assistants/AI smart technologies]<br>Using [AI virtual assistants/AI smart technologies] is something I would do in a future. | 1 (strongly disagree) – 5 (strongly agree) | $\alpha = .98$ | $\alpha = .91$ |
| Behavioral intention of users (Study 1 & Study 2) | I intend to continue using [AI virtual assistants/AI smart technologies]<br>I predict that I would continue using [AI virtual assistanst/AI smart technologies]<br>Using [AI virtual assistants/AI smart technologies] is something I would continue to do. | | $\alpha = .95$ | |
| Human-like trust in smart technologies (Study 2) | Smart technologies care about our well-being. (Benevolence)<br>Smart technologies are sincerely concerned about addressing the problems of human users. (Benevolence)<br>Smart technologies try to be helpful and do not operate out of selfish interest. (Benevolence)<br>Smart technologies are truthful in their dealings. (Integrity)<br>Smart technologies keep their commitments and deliver on their promises. (Integrity)<br>Smart technologies are honest and do not abuse the information and advantage they have over their users. (Integrity) | 1 (strongly disagree) – 5 (strongly agree) | N/A | $\alpha = .92$ |
| Functionality trust in smart technologies (Study 2) | Smart technologies work well. (Competence)<br>Smart technologies have the features necessary to complete key tasks. (Competence)<br>Smart technologies are competent in their area of expertise. (Competence) | 1 (strongly disagree) – 5 (strongly agree) | N/A | $\alpha = .91$ |

Smart technologies are reliable.
(Competence)
Smart technologies are dependable.
(Competence)

*Note.* For perceived ease of use, perceived usefulness, attitude, and behavior intention items, the same question wordings were used in Study 1 and Study 2 except the description of AI technologies. Study 1 participants were asked about their perceptions of "AI virtual assistant" and Study 2 participants were asked about "AI smart technologies."

Author Biographies

**Hyesun Choung** is a Postdoctoral Research Associate at Michigan State University in the College of Communication Arts & Sciences. She is a media effects researcher interested in social implications of emerging information technologies. Her recent research includes the automation of journalism and ethical implications of automated decision-making.

**Prabu David** is the Dean of the College of Communication Arts & Sciences at Michigan State University. He is a communication researcher who studies media and cognition. His recent research includes multitasking, mobile health, problematic use of social media, and trustworthy and ethical AI.

**Arun Ross** is the Cillag Endowed Chair in Engineering and a Professor in the Department of Computer Science and Engineering at Michigan State University. He also serves as the Site Director of the NSF Center for Identification Technology Research (CITeR). His research interests include Machine Learning, Computer Vision, and Biometrics.